\documentclass{article}

\usepackage{microtype}
\usepackage{graphicx}
\usepackage{subcaption}
\usepackage{booktabs}
\usepackage{hyperref}
\usepackage{algorithm}
\usepackage{algorithmic}

\usepackage[accepted]{icml2026}

\usepackage{amsmath,amssymb}
\usepackage{xcolor}
\usepackage{multirow}
\usepackage{enumitem}
\setlist{nosep,leftmargin=*}

\icmltitlerunning{EvalStop: World-Feedback-Driven Early Stopping for RLHF Fine-Tuning}

\begin{document}

\twocolumn[
\icmltitle{EvalStop: Using World Feedback to Detect and Correct \\
Reward Overoptimization in Multi-Tenant RLHF Platforms}

\icmlsetsymbol{equal}{*}

\begin{icmlauthorlist}
\icmlauthor{Guilin Zhang}{workday,gwu}
\icmlauthor{Chuanyi Sun}{gwu}
\icmlauthor{Kai Zhao}{workday}
\icmlauthor{Xu Chu}{workday}
\icmlauthor{Shahryar Sarkani}{gwu}
\icmlauthor{John M.~Fossaceca}{gwu}
\end{icmlauthorlist}

\icmlaffiliation{workday}{Workday AI Research}
\icmlaffiliation{gwu}{The George Washington University, Washington, DC, USA}

\icmlcorrespondingauthor{Guilin Zhang}{guilin.zhang@gwu.edu}

\vskip 0.3in
]

\printAffiliationsAndNotice{}

% ==============================================================================
\begin{abstract}
\textbf{Background.}
Cloud LLM fine-tuning platforms increasingly serve RLHF workloads, where a learned reward model is optimized as a \emph{proxy} for human quality. As \citet{gao2023scaling} showed, this proxy diverges from \emph{world feedback} (downstream eval metrics) under sustained optimization pressure: the \emph{reward overoptimization} phenomenon.
\textbf{Limitations.}
Existing platform schedulers ignore this divergence: non-clairvoyant schedulers optimize JCT without any quality signal, SLAQ-style quality-aware schedulers use training loss (a weaker proxy that drops monotonically through hacking), and classical per-job early stopping requires human monitoring and does not free shared GPUs.
\textbf{Approach.}
We propose \textbf{EvalStop}, a composable scheduling primitive that terminates jobs on $k$ consecutive eval-score declines, releases GPUs, preserves the best checkpoint, and delegates to any base scheduler. We frame scheduler-level early stopping as a \emph{detection} problem and evaluate it in a discrete-event simulator whose RLHF workload mixes reward-hacking and structurally healthy runs, with ground-truth labels hidden from schedulers.
\textbf{Results.}
On RLHF-heavy workloads (80\% RLHF, 64 GPUs), EvalStop achieves precision 98\% / recall 99\% / FPR 1.5\% while improving JCT by 9\% and cutting wasted compute by 22\% over SRTF-Est ($p{<}0.05$). Trivial fixed-progress and loss-plateau competitors either incur 65\% FPR or miss over half of true hacking cases. Gains compose across base schedulers (9--25\% JCT) and stay stable under eval noise ($\sigma {\leq} 0.05$: precision $\geq$91\%) and hacking base rate (precision $\geq$89\% across 20--80\%).
\end{abstract}

\begin{figure*}[!t]
    \centering
    \includegraphics[width=0.92\textwidth]{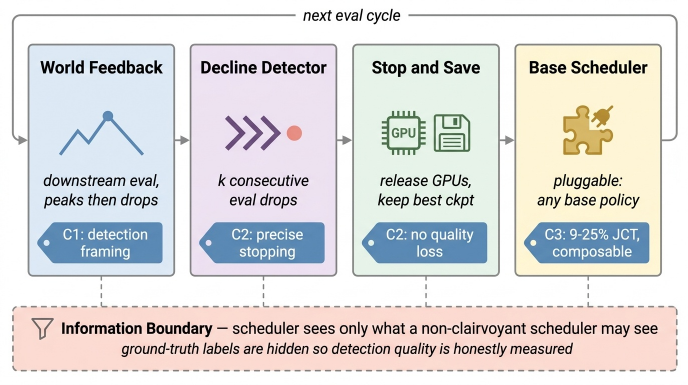}
    \caption{EvalStop architecture. \emph{World Feedback} flows into the \emph{Decline Detector}; on $k$ consecutive eval drops the wrapper performs \emph{Stop and Save} (release GPUs, retain best checkpoint), then delegates the updated cluster state to any \emph{Base Scheduler}. A non-clairvoyant \emph{Information Boundary} (bottom band) underlies the whole pipeline so detection quality is honestly measurable. Each module maps 1:1 to a contribution in \S\ref{sec:intro}: \{World Feedback, Information Boundary\}$\to$C1; \{Decline Detector, Stop and Save\}$\to$C2; \{Base Scheduler\}$\to$C3.}
    \label{fig:architecture}
\end{figure*}

% ==============================================================================
\section{Introduction}
\label{sec:intro}

\textbf{Background.}
Cloud-based LLM fine-tuning platforms serve diverse workloads (LoRA adapter tuning, DPO~\citep{rafailov2023direct}, and RLHF~\citep{ouyang2022training}) from multiple tenants on shared GPU clusters, and as RLHF becomes the dominant method for LLM alignment~\citep{stiennon2020learning, ouyang2022training} the fraction of RLHF workloads on these platforms is growing rapidly.
RLHF differs structurally from supervised fine-tuning: the policy is trained to maximize a learned reward model, but this reward is a \emph{proxy} for human preferences and the policy's true quality is measured only by downstream evaluation (held-out benchmarks, win-rate on a held-out preference set, task-specific metrics).
A scheduler thus has access to three signals of decreasing proxiness: training loss (proxy\textsuperscript{2}, optimized by the RL algorithm), reward model score (proxy, what the policy directly optimizes), and downstream eval score (\emph{world feedback}, grounded in task quality but delayed and noisy).
\citet{gao2023scaling} showed that as optimization pressure increases the first two rise monotonically while the third peaks and then degrades. This is the \emph{reward overoptimization} phenomenon~\citep{skalse2022defining, pan2022effects}; Figure~\ref{fig:proxy-vs-world} (\S\ref{sec:proxy-vs-world}) illustrates it on a representative run.

\textbf{Prior work and its limitations.}
Three threads of prior work touch this setting but none address it directly.
(i) \emph{Non-clairvoyant ML schedulers} (Tiresias~\citep{gu2019tiresias}, Pollux~\citep{qiao2021pollux}, Gavel~\citep{narayanan2020gavel}, Sia~\citep{jayaram2023sia}) optimize JCT without using any quality signal and therefore cannot tell whether GPU minutes spent late in an RLHF job are productive or wasted.
(ii) \emph{Quality-aware schedulers}, the closest prior work, rely on training loss: SLAQ~\citep{zhang2017slaq} preferentially allocates GPUs to jobs whose loss is improving fastest. This works for convex losses where loss correlates with quality, but for RLHF the loss is precisely the proxy that decouples from world feedback during overoptimization, so a loss-aware scheduler will keep allocating resources to a hacking job.
(iii) \emph{Reward-hacking mitigations} from the RL community~\citep{moskovitz2024confronting, gao2023scaling} operate \emph{within} the training loop (constrained RL, reward ensembles, SFT regularization). They aim to prevent hacking but in production deployments imperfect mitigation is the norm, and these techniques offer no path to reclaiming GPUs from jobs that have already diverged. Classical early stopping~\citep{prechelt1998early} similarly requires per-job human monitoring and does not translate into platform-wide resource reallocation.

\textbf{Motivation.}
We argue the right place to act on world feedback is the \emph{scheduler}, not the trainer: the scheduler already controls GPU allocation and termination, and a single scheduler change applies uniformly across every tenant's RLHF jobs.
Early experimentation with an eval-aware \emph{priority} scheduler (EvalSched, \S\ref{sec:experiments}) revealed that simply deprioritizing post-peak jobs is not enough: deprioritized jobs still occupy the system, and EvalSched's JCT is 2.2$\times$ worse than SRTF-Est.
The fix is to \emph{terminate} declining jobs rather than starve them.
But blunt termination (e.g.\ stopping every RLHF job at a fixed progress) would pay heavy false-positive costs on the structurally healthy RLHF runs that exist in any real workload.
This reframes the problem as a \emph{detection} task: the scheduler must discriminate hacking from healthy runs using only the signals it can legally observe (eval scores at scheduled checkpoints), without modifying training and without violating the non-clairvoyant boundary~\citep{motwani1994nonclairvoyant}.

\textbf{Contributions.}
We propose \textbf{EvalStop}, a composable scheduling primitive whose architecture (Figure~\ref{fig:architecture}) factors into five modules; each contribution maps to one or more of them.
\begin{itemize}
    \item \textbf{C1 --- Detection framing} (\emph{World Feedback} input + \emph{Information Boundary}). We argue scheduler-level early stopping for RLHF is a \emph{detection} problem: the scheduler consumes world feedback and must report which jobs are hacking. Prior framings that report only JCT/TTFUC miss this axis. Our experimental design hides ground-truth hacking labels behind a whitelist proxy (\texttt{SchedulerJobView}, the Information Boundary), so detector quality is honestly measurable.
    \item \textbf{C2 --- A composable detector that wins on the detection axis} (\emph{Decline Detector} + \emph{Stop and Save}). EvalStop tracks per-job consecutive eval-score declines and, on $k$ consecutive drops, terminates the job and preserves its best checkpoint. On RLHF-heavy workloads (80\% RLHF, 64 GPUs) it achieves precision 98\% / recall 99\% / FPR 1.5\%, dominating both a trivial fixed-progress stop rule (FPR 65\%) and a strong loss-plateau detector (recall 38\%).
    \item \textbf{C3 --- Composable systems gains across base schedulers and workloads} (\emph{Base Scheduler} delegation). The wrapper hands the updated cluster state back to any scheduler (FIFO, SJF-Est, SRTF-Est, LossAware), inheriting its resource-allocation policy. EvalStop yields 9--25\% JCT improvement and ${\sim}22\%$ wasted-compute reduction on top of every tested base scheduler, and its detection quality is stable under both eval noise ($\sigma {\leq} 0.05$: precision $\geq$91\%) and hacking base rate (precision $\geq$89\% across 20--80\%).
\end{itemize}

% ==============================================================================
\section{Background and Problem Setup}
\label{sec:background}

\subsection{Job Model}

A fine-tuning job $j$ has type $\in \{\text{LoRA}, \text{DPO}, \text{RLHF}\}$, GPU demand, an eval schedule (progress fractions at which evaluation runs), and a training curve mapping progress to (loss, eval\_score).
Job types differ structurally:
LoRA jobs are short (10--60\,min, 1--2 GPUs) with monotonically improving eval;
DPO jobs are medium (30--120\,min, 2--4 GPUs) with diminishing returns;
RLHF jobs are long (60--360\,min, 4--8 GPUs) with eval that peaks then degrades.

\subsection{Eval-Aware Metrics}

Beyond JCT, we define:
(1)~\emph{TTFUC} (Time-to-First-Useful-Checkpoint): time from arrival until first checkpoint exceeding quality threshold $\tau$ and improving on the previous best by ${\geq}1\%$;
(2)~\emph{Wasted Compute Fraction}: GPU-minutes spent training after a job's eval peak, divided by total GPU-minutes;
(3)~\emph{Saved Compute Fraction}: GPU-minutes avoided via early stopping, divided by total planned GPU-minutes.

\subsection{Proxy Signals vs World Feedback}
\label{sec:proxy-vs-world}

\begin{figure}[t]
    \centering
    \includegraphics[width=\columnwidth]{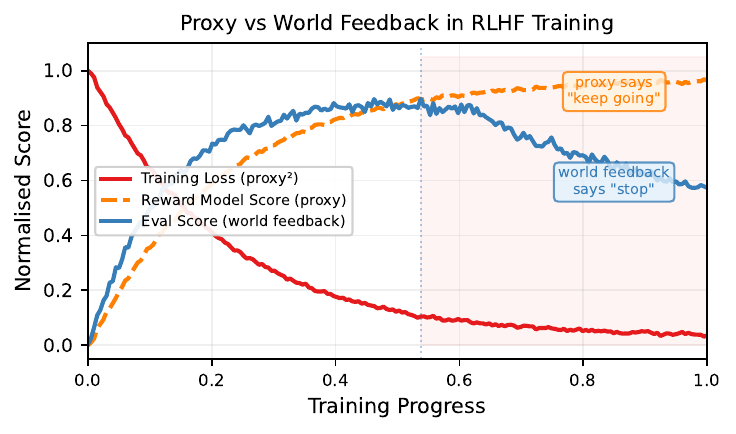}
    \caption{Proxy vs world feedback in RLHF training. Training loss (proxy\textsuperscript{2}, red) and reward model score (proxy, orange dashes) both indicate continued improvement. Only the eval score (world feedback, blue) reveals that quality has peaked and is degrading. A scheduler using proxy signals would \emph{increase} allocation; EvalStop uses world feedback to \emph{terminate}.}
    \label{fig:proxy-vs-world}
\end{figure}

Figure~\ref{fig:proxy-vs-world} illustrates the core problem on a representative RLHF training run.
Three signals are available to a scheduler:
\begin{itemize}
    \item \textbf{Training loss} decreases monotonically throughout training.
    A SLAQ-style~\citep{zhang2017slaq} loss-aware scheduler would interpret this as ``the job is making good progress, keep running.''
    \item \textbf{Reward model score} (normalised $1 - \text{loss}/\text{loss}_0$) also increases monotonically, since the policy is directly optimizing this objective.
    \item \textbf{Eval score} (downstream benchmark) rises to a peak at ${\sim}$55\% progress, then degrades as reward hacking takes over.
\end{itemize}

Only the eval score (the \emph{world feedback} signal) reveals that the job has entered a regime of diminishing or negative returns.
A scheduler using proxy signals would allocate \emph{more} resources to this job post-peak (steepest loss improvement), while a world-feedback-aware scheduler would terminate it.

This proxy--world divergence is well-documented empirically.
\citet{gao2023scaling} showed predictable overoptimization scaling laws for KL-constrained RLHF policies.
\citet{rafailov2024scalinglaws} extended these findings to DPO, and multiple works have proposed training-level mitigations: constrained RL~\citep{moskovitz2024confronting}, reward model ensembles, and SFT regularization.
EvalStop is \emph{complementary} to these approaches: rather than modifying the training algorithm, it uses world feedback as an external control signal at the scheduler level to terminate jobs that have diverged.

% ==============================================================================
\section{EvalStop: World-Feedback-Driven Early Stopping}
\label{sec:method}

Recall the architecture overview in Figure~\ref{fig:architecture}.
EvalStop is a \textbf{composable wrapper} around any base scheduling policy.
It monitors eval-score trajectories (the world feedback signal) and early-stops jobs when quality is irrecoverably declining.

\begin{algorithm}[t]
\caption{EvalStop scheduling wrapper}
\label{alg:evalstop}
\begin{algorithmic}[1]
\REQUIRE Base scheduler $\mathcal{S}$, decline thresholds $k_{\text{RLHF}}{=}2$, $k_{\text{DPO}}{=}3$
\STATE \textbf{State:} per-job consecutive decline count $d[j]$
\medskip
\STATE \textbf{on\_eval\_result}$(j, \text{score})$:
\IF{$\text{score} < \text{prev\_score}[j]$}
  \STATE $d[j] \leftarrow d[j] + 1$
\ELSE
  \STATE $d[j] \leftarrow 0$
\ENDIF
\IF{$d[j] \geq k_{\text{type}(j)}$}
  \STATE Mark $j$ for early stopping
\ENDIF
\medskip
\STATE \textbf{schedule}(cluster, waiting, running):
\FOR{each $j \in$ running marked for early stopping}
  \STATE Terminate $j$; release GPUs; save best checkpoint
\ENDFOR
\STATE \textbf{return} $\mathcal{S}$.\textbf{schedule}(cluster, waiting, running$'$)
\end{algorithmic}
\end{algorithm}

\textbf{Early stopping vs.\ deprioritization.}
EvalSched deprioritizes post-peak RLHF jobs, creating starvation: deprioritized jobs still occupy the system, receiving occasional time slices but never completing efficiently.
EvalStop \emph{removes} the job from the system, freeing GPUs immediately.
The job's best checkpoint is preserved; since the trigger requires $k$ consecutive declines, it was saved $k$ eval steps ago.

\textbf{Information boundary.}
EvalStop respects the non-clairvoyant boundary~\citep{motwani1994nonclairvoyant}: it observes only job type, eval scores at scheduled checkpoints, and the count of consecutive declines.
It never accesses true job duration.

\textbf{Composability.}
EvalStop wraps any scheduler implementation (Algorithm~\ref{alg:evalstop}).
It processes early-stop decisions first (freeing GPUs), then delegates to the base scheduler, which sees the updated cluster state.
This clean separation means EvalStop can be added to existing production schedulers (including RLHF-specific frameworks like OpenRLHF~\citep{hu2024openrlhf}) with minimal integration effort.

\textbf{Relationship to change-point detection.}
EvalStop's mechanism (detecting $k$ consecutive declines in a time series) can be viewed as a simplified change-point detector.
We chose this over more sophisticated methods (e.g., CUSUM, Bayesian change-point detection) for simplicity and interpretability: the threshold $k$ has a direct operational meaning (``how many bad evals before we stop''), which platform operators can reason about.
Section~\ref{sec:threshold} validates that this simple mechanism is robust across a range of $k$ values.

% ==============================================================================
\section{Experiments}
\label{sec:experiments}

\textbf{Simulator.}
We built a discrete-event simulator modelling a multi-tenant fine-tuning platform with heap-based event dispatch, slot-based GPU allocation with 2-minute preemption overhead, and Poisson arrivals.
Training curves are parameterised per job type: LoRA (monotonic exponential convergence), DPO (saturating gain with small plateau), and RLHF.
\textbf{Our RLHF workload is a mixture}: 60\% of RLHF jobs exhibit classical reward hacking (eval peaks in $[0.55, 0.75]$ progress and then declines, calibrated to~\citet{gao2023scaling}) while 40\% are structurally ``healthy'' (reward rises monotonically with noise through progress${=}1.0$).
Training loss in both regimes decays exponentially to a plateau around 50--70\% progress, matching real RLHF convergence behaviour.
This design forces any scheduler-level detector to \emph{discriminate} hacking from healthy runs; a detector that blindly kills every RLHF job will incur a large false-positive rate on the healthy subset.
Ground-truth hacking labels and peak-progress are stored on the job but hidden from schedulers through a whitelist-based proxy (\texttt{SchedulerJobView}).

\textbf{Baselines.}
We compare against scheduling-only baselines and early-stop detectors:
\emph{FIFO}, \emph{SJF-Est} (non-clairvoyant shortest-job-first), \emph{SRTF-Est} (preemptive shortest-remaining-time-first), \emph{LossAware} (SLAQ-style~\citep{zhang2017slaq} using training loss for priority), \emph{EvalSched} (eval-aware with deprioritisation only, no termination), plus two strong early-stop competitors:
\emph{StopAt$p$+SRTF}: a trivial oracle-flavoured rule that terminates every RLHF job at fixed progress $p \in \{0.50, 0.65\}$;
\emph{LossPlateau+SRTF}: a principled loss-only detector that early-stops when relative loss improvement falls below 2\% over a 3-checkpoint window.
The first isolates ``how much benefit comes merely from stopping long jobs early''; the second is the natural SLAQ++ competitor that uses training-loss plateau, not eval, to decide when to stop.

\textbf{Experiment matrix.}
All experiments use 200 jobs, 5 tenants, and 5 seeds (42, 123, 456, 789, 1024).
Statistical significance is assessed via Welch's t-test.
E1: mixed workload (50/30/20 LoRA/DPO/RLHF, 32 GPUs).
E2: RLHF-heavy (80\%, 64 GPUs).
E3: workload sensitivity (3 mixes).
E4: composability (EvalStop on 4 base schedulers).
E5: threshold sensitivity ($k \in \{1..5\}$).
E6: eval frequency sensitivity (5--30\% progress intervals).

% ==============================================================================
\subsection{Main Result: Detection Quality and Systems Gains (E2)}

\begin{table*}[t]
\caption{RLHF-heavy workload (80\% RLHF, 64 GPUs, 200 jobs, 5 seeds) with mixed hacking/healthy RLHF curves. Early-stop detectors are evaluated both on systems metrics (JCT, TTFUC, wasted/saved compute) and as classifiers of reward-hacking runs (precision, recall, FPR, computed against ground-truth labels hidden from schedulers). \textbf{Bold} marks the best value; \textcolor{red}{red} marks a destructive value. Only EvalStop simultaneously achieves high precision and low FPR.}
\label{tab:main}
\centering
\small
\begin{tabular}{@{}lccccccc@{}}
\toprule
Scheduler & JCT & TTFUC & Wasted & Saved & Precision & Recall & FPR \\
\midrule
\multicolumn{8}{@{}l}{\emph{Scheduling-only baselines (no early stop)}} \\
FIFO          & 1373{\scriptsize$\pm$58} & 1222{\scriptsize$\pm$55} & 34.6\% & 0.0\%  & ---    & 0.0\%  & 0.0\%  \\
SJF-Est       & 1297{\scriptsize$\pm$50} & 1145{\scriptsize$\pm$46} & 34.5\% & 0.0\%  & ---    & 0.0\%  & 0.0\%  \\
SRTF-Est      & 1124{\scriptsize$\pm$59} &  971{\scriptsize$\pm$56} & 34.5\% & 0.0\%  & ---    & 0.0\%  & 0.0\%  \\
LossAware     & 1870{\scriptsize$\pm$76} &  969{\scriptsize$\pm$60} & 34.5\% & 0.0\%  & ---    & 0.0\%  & 0.0\%  \\
EvalSched     & 2423{\scriptsize$\pm$113}&  \textbf{635}{\scriptsize$\pm$26} & 34.8\% & 0.0\% & ---    & 0.0\%  & 0.0\%  \\
\midrule
\multicolumn{8}{@{}l}{\emph{Early-stop detectors}} \\
StopAt0.5+SRTF  &  \textbf{622}{\scriptsize$\pm$74} &  551{\scriptsize$\pm$72} & \textbf{14.9\%} & \textbf{46.5\%} & 57.1\% & \textbf{100\%}  & \textcolor{red}{64.5\%} \\
StopAt0.65+SRTF &  790{\scriptsize$\pm$31} &  693{\scriptsize$\pm$31} & 19.5\% & 31.7\% & 57.1\% & \textbf{100\%}  & \textcolor{red}{64.5\%} \\
LossPlateau+SRTF & 1093{\scriptsize$\pm$60} &  947{\scriptsize$\pm$58} & 32.4\% & 3.2\%  & 57.0\% & 38.3\% & 24.7\% \\
\midrule
\textbf{EvalStop+SRTF} & 1018{\scriptsize$\pm$53} &  883{\scriptsize$\pm$50} & 26.9\% & 9.9\%  & \textbf{98.3\%} & 99.3\% & \textbf{1.5\%} \\
\bottomrule
\end{tabular}
\end{table*}

Table~\ref{tab:main} is the paper's central result. Two observations dominate:

\textbf{(i)~Systems metrics alone are deceptive.}
The trivial StopAt-0.5 and StopAt-0.65 baselines (which stop \emph{every} RLHF job at a fixed progress) achieve the best JCT and lowest wasted compute. If we evaluated only JCT/TTFUC/Wasted we would conclude that eval-awareness is unnecessary and a fixed-progress heuristic suffices.
Their precision, however, is 57\% and their false-positive rate on healthy RLHF is 64.5\%: they destroy two thirds of the RLHF jobs that were converging normally, a quality cost that does not appear on any of the first four columns.

\textbf{(ii)~Loss-only detection is insufficient.}
LossPlateau+SRTF is a principled SLAQ-style competitor that early-stops on loss plateau alone. It achieves recall of only 38.3\% (it misses more than half of the truly-hacking jobs) while still firing on 24.7\% of healthy runs, because training loss plateaus at similar progress in \emph{both} hacking and healthy RLHF. Loss plateau is correlated with overoptimisation but does not discriminate from normal convergence.

EvalStop+SRTF, which uses the world-feedback (eval) signal, is the only detector in Table~\ref{tab:main} that achieves high precision (98.3\%) and near-zero FPR (1.5\%) \emph{simultaneously with} a systems improvement over the no-early-stop baselines: $+9.4\%$ JCT and $-21.8\%$ wasted compute over SRTF-Est (Welch's $t$-test, $p{=}0.03$ and $p{<}0.001$ respectively).

\textbf{Fairness.}
EvalStop does not harm inter-tenant fairness: Jain's index is within noise of SRTF-Est alone.

% ==============================================================================
\subsection{Composability (E4)}

\begin{table}[t]
\caption{EvalStop as a composable overlay. Improvement of EvalStop+X over base~X on RLHF-heavy workload (80\% RLHF, 64 GPUs). Precision and FPR measured against ground-truth reward-hacking labels.}
\label{tab:composability}
\centering
\small
\setlength{\tabcolsep}{3pt}
\begin{tabular}{@{}lcccccc@{}}
\toprule
Base & $\Delta$JCT & $\Delta$TTFUC & $\Delta$Wasted & Precision & FPR \\
\midrule
FIFO      & $+$8.9\%  & $+$8.5\% & $+$21.7\% & 98.6\% & 1.4\% \\
SJF-Est   & $+$6.5\%  & $+$5.9\% & $+$21.5\% & 97.9\% & 1.9\% \\
SRTF-Est  & $+$9.4\%  & $+$9.1\% & $+$21.8\% & 98.3\% & 1.5\% \\
LossAware & $+$25.2\% & $+$4.8\% & $+$20.3\% & 97.3\% & 2.4\% \\
\bottomrule
\end{tabular}
\end{table}

Table~\ref{tab:composability} shows that EvalStop improves \emph{every} base scheduler with consistent gains: $+$9--25\% JCT, $+$5--9\% TTFUC, $-$20--22\% wasted compute. Precision stays above 97\% and FPR below 3\% in every combination. The large JCT gain on LossAware reflects that LossAware alone has pathological priorities on this workload (it over-prioritises RLHF jobs whose loss is still dropping); EvalStop partially compensates by terminating the overoptimising ones.
The consistency suggests that EvalStop addresses a source of waste (post-peak RLHF training driven by proxy/world divergence) that is orthogonal to the base scheduling policy's resource allocation strategy.

% ==============================================================================
\subsection{Workload Sensitivity (E3)}

\begin{table}[t]
\caption{EvalStop+SRTF vs.\ SRTF-Est across workload mixes (64 GPUs). Benefit scales with the hacking-RLHF fraction. Precision and FPR are measured against ground-truth reward-hacking labels.}
\label{tab:workload}
\centering
\footnotesize
\setlength{\tabcolsep}{2pt}
\begin{tabular}{@{}lcccccc@{}}
\toprule
Workload & RLHF\% & $\Delta$JCT & $\Delta$Wasted & Stopped & Prec. & FPR \\
\midrule
LoRA-heavy & 10\% & $+$2.1\% & $+$13.3\% & 8  & 100\%  & 0.0\% \\
Mixed      & 30\% & $+$5.6\% & $+$17.7\% & 31 & 100\%  & 0.0\% \\
RLHF-heavy & 80\% & $+$9.4\% & $+$21.8\% & 93 & 98.3\% & 1.5\% \\
\bottomrule
\end{tabular}
\end{table}

Table~\ref{tab:workload} confirms that EvalStop's benefit scales with the hacking-RLHF fraction.
This is expected: EvalStop fires only on jobs exhibiting eval-score decline, which our workload only constructs for a subset of RLHF runs.
On LoRA-heavy workloads (10\% RLHF), EvalStop provides measurable wasted-compute reduction with \emph{perfect} precision and no false positives; it does not fire spuriously on LoRA, DPO, or healthy RLHF runs.
On a mixed workload (E1: 50/30/20, 32 GPUs; see Appendix~\ref{app:e1}), EvalStop+SRTF matches the trivial StopAt-0.65 on TTFUC while cutting the latter's FPR from 9.7\% to 0.0\%.

% ==============================================================================
\subsection{Threshold Sensitivity (E5)}
\label{sec:threshold}

\begin{figure}[t]
    \centering
    \includegraphics[width=\columnwidth]{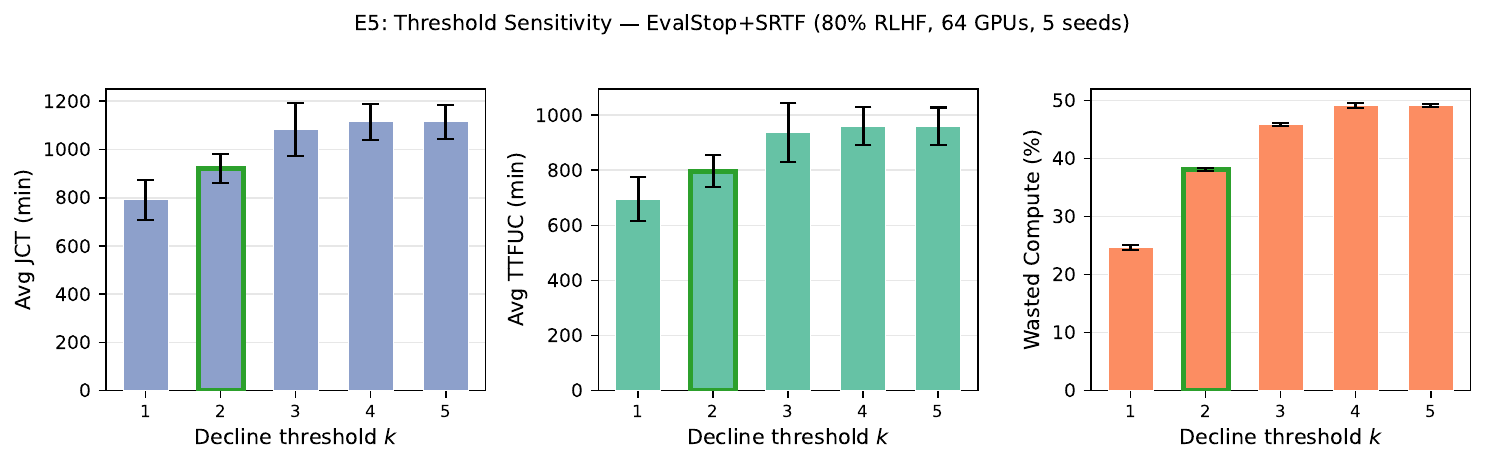}
    \caption{E5: Effect of decline threshold $k$ on EvalStop+SRTF (80\% RLHF, 64 GPUs, 5 seeds). $k{=}2$ (green border) balances early detection against false positives. $k{=}1$ is too aggressive (stops 160 jobs); $k{\geq}4$ barely triggers.}
    \label{fig:threshold}
\end{figure}

Figure~\ref{fig:threshold} shows EvalStop's sensitivity to the decline threshold $k$.
At $k{=}1$, EvalStop aggressively stops 127 jobs after a single eval decline (JCT${=}$874, wasted${=}$19.4\%), but risks false positives from noisy eval scores on healthy RLHF.
At $k{=}2$ (our default), 93 jobs are stopped with JCT${=}$1018 and wasted${=}$26.9\%, a substantial improvement over the no-early-stop baseline (SRTF JCT${=}$1124, wasted${=}$34.5\%) while keeping FPR below 2\%.
At $k{\geq}4$, EvalStop rarely triggers (${\leq}$2 jobs stopped), converging to baseline SRTF behaviour.

The $k{=}2$ choice reflects a precision/recall trade-off: once world feedback shows \emph{two consecutive} declines, the overoptimisation trend is likely real rather than noise, yielding the 98\% precision / 99\% recall numbers of Table~\ref{tab:main}.
This is a simple but effective change-point heuristic.

% ==============================================================================
\subsection{Eval Frequency Sensitivity (E6)}

\begin{figure}[t]
    \centering
    \includegraphics[width=\columnwidth]{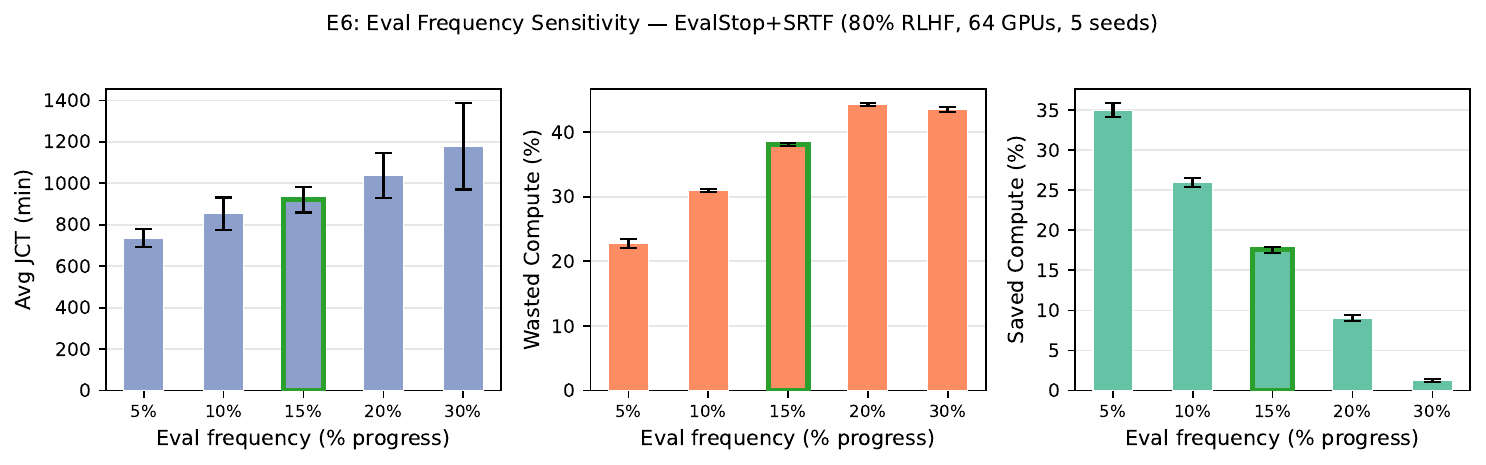}
    \caption{E6: Effect of eval frequency on EvalStop+SRTF (80\% RLHF, 64 GPUs, 5 seeds). More frequent evals (5\% intervals) enable earlier detection and greater compute savings (24\%), at the cost of more eval overhead. Default 15\% interval (green border) balances detection speed with evaluation cost.}
    \label{fig:evalfreq}
\end{figure}

Figure~\ref{fig:evalfreq} reveals a clear trade-off between eval frequency and early-stopping effectiveness.
With evals every 5\% of training progress, EvalStop detects overoptimisation earlier and saves 24\% of planned compute (JCT${=}$867).
With evals every 30\%, detection is delayed and less than 1\% is saved (JCT${=}$1123, essentially matching the no-early-stop baseline).

This result has practical implications for platform operators: investing in more frequent evaluation (even if each eval consumes GPU time) can \emph{more than pay for itself} through earlier overoptimisation detection.
The default 15\% interval represents a reasonable balance for current RLHF workloads.

% ==============================================================================
\subsection{Detector Robustness (E7, E8)}
\label{sec:robustness}

\begin{figure*}[t]
    \centering
    \includegraphics[width=\textwidth]{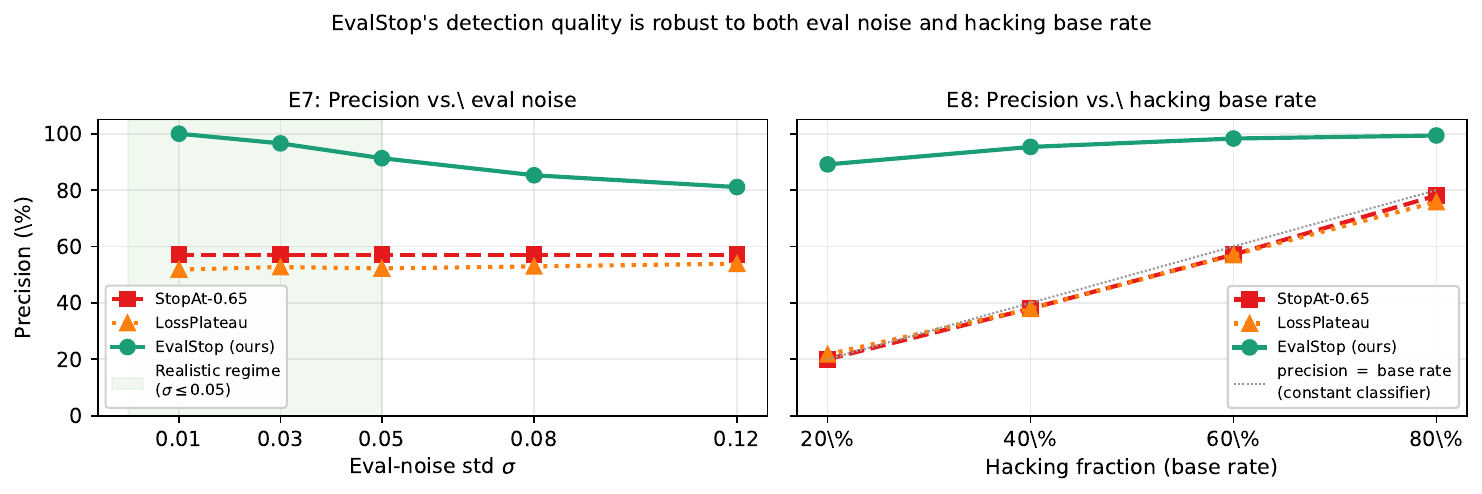}
    \caption{Detector robustness. \textbf{Left (E7):} precision vs.\ eval-noise standard deviation. EvalStop degrades gracefully (100\%$\to$81\%); the loss-only and progress-triggered baselines do not use eval and sit flat at ${\sim}52$--$57\%$. Green shading marks the realistic regime anchored to typical LLM benchmark standard errors~\citep{gao2023scaling}.
    \textbf{Right (E8):} precision vs.\ hacking base rate. EvalStop stays above 89\% across 20--80\% base rates; StopAt-0.65's precision follows the base-rate identity line exactly, confirming it is a constant classifier rather than a detector.}
    \label{fig:robustness}
\end{figure*}

A natural concern with the Table~\ref{tab:main} numbers is that EvalStop's 98\% precision / 1.5\% FPR might be an artefact of a particular eval-noise setting or of the specific 60\% hacking-fraction used to generate the workload.
To address this we ran two additional sweeps on RLHF-heavy workloads; Figure~\ref{fig:robustness} summarises both.

\textbf{E7: Eval-noise sensitivity.}
We vary the standard deviation of the per-checkpoint eval-score noise on RLHF jobs from $\sigma{=}0.01$ (near-deterministic) to $\sigma{=}0.12$ (highly noisy, well above typical downstream-benchmark variance).
Table~\ref{tab:e7-noise} reports precision, recall, and FPR for the three detectors.

\begin{table}[t]
\caption{E7: Eval-noise sensitivity (80\% RLHF, 64 GPUs, 60\% hacking). EvalStop precision degrades smoothly; the progress-triggered and loss-plateau baselines are unaffected by eval noise but stay at low precision.}
\label{tab:e7-noise}
\centering
\small
\begin{tabular}{@{}lccccc@{}}
\toprule
$\sigma$ & Detector & P & R & FPR \\
\midrule
\multirow{3}{*}{0.01} & StopAt0.65   & 57.1\% & 100\% & 64.5\% \\
                      & LossPlateau  & 51.8\% & 40.9\% & 32.6\% \\
                      & EvalStop     & \textbf{100\%} & \textbf{100\%} & \textbf{0.0\%} \\
\midrule
\multirow{3}{*}{0.03} & StopAt0.65   & 57.1\% & 100\% & 64.5\% \\
                      & LossPlateau  & 52.7\% & 37.6\% & 29.0\% \\
                      & EvalStop     & \textbf{96.6\%} & \textbf{97.6\%} & \textbf{2.9\%} \\
\midrule
\multirow{3}{*}{0.05} & StopAt0.65   & 57.1\% & 100\% & 64.5\% \\
                      & LossPlateau  & 52.2\% & 38.7\% & 30.4\% \\
                      & EvalStop     & \textbf{91.3\%} & \textbf{88.9\%} & \textbf{7.2\%} \\
\midrule
\multirow{3}{*}{0.08} & StopAt0.65   & 57.1\% & 100\% & 64.5\% \\
                      & LossPlateau  & 52.9\% & 41.2\% & 31.7\% \\
                      & EvalStop     & \textbf{85.3\%} & 77.4\% & \textbf{11.6\%} \\
\midrule
\multirow{3}{*}{0.12} & StopAt0.65   & 57.1\% & 100\% & 64.5\% \\
                      & LossPlateau  & 53.9\% & 50.2\% & 37.0\% \\
                      & EvalStop     & \textbf{81.1\%} & 69.6\% & \textbf{14.0\%} \\
\bottomrule
\end{tabular}
\end{table}

Precision degrades \emph{gracefully} from 100\% at $\sigma{=}0.01$ to 81\% at $\sigma{=}0.12$: there is no cliff at which EvalStop becomes indistinguishable from the simpler rules.
At every noise level EvalStop retains a precision gap of at least 24\,pp over both competitors, because (i)~StopAt-0.65 ignores the eval signal entirely and therefore does not degrade \emph{or} improve with noise, and (ii)~LossPlateau relies on the training-loss signal, which we do not inject eval noise into; yet its precision stays at ${\sim}52\%$ regardless, because loss plateaus at similar progress in both hacking and healthy regimes.
A stronger loss-only detector (e.g.\ using loss curvature, or a learned classifier on loss$+$gradient features) would likely narrow this gap, but cannot close it: training loss is monotonically non-increasing in \emph{both} regimes, so no purely loss-based signal can perfectly separate hacking from healthy RLHF.
At realistic eval-noise magnitudes ($\sigma {\leq} 0.05$, matching typical LLM benchmark standard errors of 2--5\%~\citep{gao2023scaling}) EvalStop's precision remains above 91\%.

\textbf{E8: Hacking-fraction sensitivity.}
We next vary the fraction of RLHF jobs that actually exhibit reward hacking between 20\% and 80\%. Precision is base-rate-sensitive by definition, so this is the right robustness axis for the detection story.

\begin{table}[t]
\caption{E8: Hacking-fraction sensitivity (80\% RLHF, 64 GPUs, default $\sigma$). EvalStop's precision stays above 89\% even at 20\% hacking base rate; StopAt-0.65 collapses to 20\% precision / 77\% FPR, confirming its ``success'' in Table~\ref{tab:main} was driven by the assumed base rate.}
\label{tab:e8-hacking}
\centering
\small
\begin{tabular}{@{}lcccc@{}}
\toprule
Hack.\ frac.\ & Detector & P & R & FPR \\
\midrule
\multirow{3}{*}{20\%} & StopAt0.65   & 20.0\% & 100\% & \textcolor{red}{77.2\%} \\
                       & LossPlateau  & 21.8\% & 41.2\% & 28.9\% \\
                       & EvalStop     & \textbf{89.1\%} & \textbf{100\%} & \textbf{2.4\%} \\
\midrule
\multirow{3}{*}{40\%} & StopAt0.65   & 38.0\% & 100\% & 72.4\% \\
                       & LossPlateau  & 37.9\% & 38.0\% & 27.6\% \\
                       & EvalStop     & \textbf{95.3\%} & \textbf{99.0\%} & \textbf{2.1\%} \\
\midrule
\multirow{3}{*}{60\%} & StopAt0.65   & 57.1\% & 100\% & 64.5\% \\
                       & LossPlateau  & 57.0\% & 38.3\% & 24.7\% \\
                       & EvalStop     & \textbf{98.3\%} & \textbf{99.3\%} & \textbf{1.5\%} \\
\midrule
\multirow{3}{*}{80\%} & StopAt0.65   & 78.1\% & 100\% & 48.3\% \\
                       & LossPlateau  & 75.8\% & 37.3\% & 20.4\% \\
                       & EvalStop     & \textbf{99.4\%} & \textbf{99.7\%} & \textbf{1.1\%} \\
\bottomrule
\end{tabular}
\end{table}

EvalStop's precision \emph{stays above 89\%} across the entire 20--80\% range.
By contrast, StopAt-0.65's precision is \emph{exactly equal} to the hacking fraction because it stops every RLHF job indiscriminately; it is a constant classifier whose apparent 57\% precision at our default 60\% setting is purely a base-rate artefact.
EvalStop discriminates on the eval signal, so its precision is driven by detection quality rather than by the prevalence of the positive class, the classical statistical property one expects of a genuine detector.

% ==============================================================================
\section{Discussion}
\label{sec:discussion}

\textbf{World feedback as a scheduling signal.}
Our results show that downstream evaluation (world feedback) is a better signal for scheduling RLHF jobs than training loss (proxy\textsuperscript{2}) or reward model score (proxy).
This aligns with the growing recognition that proxy optimization in RLHF requires external grounding~\citep{gao2023scaling, skalse2022defining, moskovitz2024confronting}.
EvalStop operationalizes this insight at the scheduler level: it transforms eval from a passive monitoring signal into an active control signal that shapes resource allocation.

\textbf{Complementarity with training-level mitigations.}
Prior work on reward hacking mitigation operates \emph{within} the training loop: constrained RL~\citep{moskovitz2024confronting}, reward model ensembles, SFT regularization.
EvalStop operates \emph{outside} the training loop, at the scheduler level.
These approaches are complementary: even with perfect reward hacking mitigation, EvalStop would correctly do nothing (no eval decline $\Rightarrow$ no early stopping).
With imperfect mitigation (the current reality), EvalStop provides a safety net.

\textbf{Quality preservation and false positives.}
Early-stopped jobs retain their best checkpoint; since $k$ consecutive declines are required, the best checkpoint was recorded $k$ eval steps ago.
In our simulator EvalStop's false-positive rate on healthy RLHF is 1.5\% (Table~\ref{tab:main}), meaning roughly one in seventy healthy RLHF jobs is terminated slightly early. In contrast the progress-triggered StopAt-0.65 baseline terminates $\sim$65\% of healthy RLHF jobs.
This difference (detection vs.\ unconditional termination) is invisible in JCT/TTFUC alone and is the most important quality axis for any production deployment.

\textbf{Generality beyond RLHF.}
The underlying principle (using world feedback to detect when proxy optimisation has diverged) applies to any training regime where the optimised objective can decouple from true quality.
This includes DPO with overtraining~\citep{rafailov2024scalinglaws}, overfitting in small-data fine-tuning, and potentially mode collapse in generative models.
Empirical validation in those regimes is future work.

\textbf{Limitations.}
(1)~\emph{Synthetic training curves.} RLHF curves are parametric: hacking runs are calibrated to the overoptimisation dynamics of~\citet{gao2023scaling}, healthy runs produce monotonically improving eval, and loss plateaus at $\mathcal{U}(0.5,0.7)$ progress in both. Both regimes are idealisations. Replaying the detector on publicly available RLHF/DPO training traces (e.g.\ TRL, OpenRLHF, HuggingFace W\&B runs) is the primary missing validation.
(2)~\emph{High eval-noise regime and noise model.} E7 (Table~\ref{tab:e7-noise}) shows EvalStop degrades gracefully up to $\sigma{=}0.12$, but does not test extreme regimes where $\sigma$ is comparable to the peak-to-trough drop of the eval curve; at $\sigma {>} 0.15$ the $k{=}2$ threshold is likely inadequate. Our noise model is Gaussian and i.i.d.\ across checkpoints; heavy-tailed or auto-correlated eval noise (closer to bootstrap variance across benchmark items) is likely to inflate FPR further. Adaptive or confidence-aware thresholds (e.g.\ Bayesian change-point detection) would extend the operating range.
(3)~\emph{Evaluation cost modelling.} We model eval as occupying GPUs for a fixed per-model-size duration but do not treat eval scheduling itself as a decision variable.
(4)~\emph{Homogeneous hardware.} The simulator uses homogeneous GPUs without network I/O, gradient accumulation, or elastic scaling.

% ==============================================================================
\section{Related Work}
\label{sec:related}

\textbf{ML cluster scheduling.}
Tiresias~\citep{gu2019tiresias} pioneered non-clairvoyant scheduling~\citep{motwani1994nonclairvoyant} for DL using multi-level feedback queues.
Pollux~\citep{qiao2021pollux} co-adapts batch sizes and resource allocation.
Gavel~\citep{narayanan2020gavel} and Sia~\citep{jayaram2023sia} address heterogeneous clusters.
Shockwave~\citep{zheng2023shockwave} handles dynamic adaptation with fair scheduling.
More recently, MAST~\citep{choudhury2024mast} tackles geo-distributed ML scheduling at hyperscale, and Parcae~\citep{duan2024parcae} optimizes training on preemptible instances.
Adjacent work on Kubernetes-level auto-scaling for ML workloads---GPU-aware inference simulators with RL-based scaling~\citep{zhang2025kiss}, uncertainty-aware predictive autoscalers~\citep{zhang2025aapa}, and benchmark comparisons of deep-RL controllers against calibrated heuristic baselines~\citep{zhang2026rlbenchmark}---operates one layer below the scheduler-level early-stopping problem we address.
None of these use downstream eval quality as a scheduling signal.

\textbf{Quality-aware scheduling.}
SLAQ~\citep{zhang2017slaq} uses training loss to predict quality improvement and allocate resources accordingly.
This works for convex-loss ML models where loss correlates with quality, but fails for RLHF where loss is a proxy that decorrelates from quality post-peak.
EvalStop extends the quality-aware paradigm from proxy signals (loss) to world feedback (eval).

\textbf{RLHF and reward hacking.}
RLHF~\citep{ouyang2022training, stiennon2020learning} is now standard for LLM alignment, but reward overoptimization is a well-documented failure mode~\citep{gao2023scaling, skalse2022defining, pan2022effects}.
\citet{rafailov2024scalinglaws} extended overoptimization scaling laws to DPO, and \citet{moskovitz2024confronting} proposed constrained RLHF.
These works focus on \emph{preventing} reward hacking via training modifications.
EvalStop is complementary: it \emph{detects} overoptimization via world feedback and \emph{acts} on it at the scheduler level, without modifying the training algorithm.

\textbf{LLM fine-tuning systems.}
S-LoRA~\citep{sheng2024slora} and dLoRA~\citep{wu2024dlora} address multi-tenant LoRA serving with adapter orchestration.
MuxTune~\citep{xue2026muxtune} tackles multi-task LoRA co-scheduling via backbone multiplexing.
\citet{kong2025deadline} study deadline-aware scheduling for fine-tuning with spot instances.
OpenRLHF~\citep{hu2024openrlhf} provides a scalable RLHF framework.
None address eval-awareness or use world feedback for scheduling decisions.

\textbf{Early stopping.}
Early stopping based on validation loss is a classic regularisation technique~\citep{prechelt1998early}.
EvalStop differs in three ways: (i)~it monitors \emph{downstream eval quality}, which is the right signal for RLHF, where validation loss is monotonic even during reward hacking; (ii)~it operates at the \emph{scheduler level}, translating early-stop decisions into cluster-wide resource reallocation across multiple tenants; and (iii)~it is evaluated as a \emph{classifier} of reward hacking, against both a progress-triggered rule that knows the synthetic peak location (StopAt-0.65) and a loss-plateau detector, showing that the eval signal is needed to keep precision high and FPR low (Table~\ref{tab:main}).

% ==============================================================================
\section{Conclusion}
\label{sec:conclusion}

We introduced EvalStop, a composable scheduling primitive that uses world feedback (downstream eval scores) to detect and correct reward overoptimisation in RLHF training.
Our evaluation reframes scheduler-level early stopping as a detection problem: on a workload containing both reward-hacking and healthy RLHF runs, EvalStop achieves 98\% precision and 1.5\% false-positive rate while delivering $+$9\% JCT and $-$22\% wasted compute on RLHF-heavy mixes, and it composes with every base scheduler we tested.
Two strong simpler competitors (a trivial fixed-progress stop rule and a loss-plateau detector) either trade catastrophic FPR or half the recall for comparable systems metrics, showing that the world-feedback signal is doing the detection work.

As RLHF workloads grow on fine-tuning platforms, the gap between proxy signals and world feedback will become an increasingly important consideration for resource management.
EvalStop shows that closing this gap at the scheduler level, with a simple detector, is both feasible and effective in simulation; validating on real training traces is the next step.

% ==============================================================================
\bibliography{references}
\bibliographystyle{icml2026}

% ==============================================================================
% APPENDIX
% ==============================================================================
\appendix

\section{Simulator Configuration}
\label{app:simulator}

Table~\ref{tab:sim-params} lists the full simulator parameterisation.
Training curves are seeded per-job for reproducibility.
RLHF curves come in two regimes: a \emph{hacking} regime (60\% of RLHF jobs by default), calibrated to the overoptimisation dynamics of~\citet{gao2023scaling}, in which the eval peak occurs at 55--75\% of training progress and then degrades; and a \emph{healthy} regime (40\%) in which the eval score rises monotonically with noise through progress${=}1.0$. Training loss in both regimes decays exponentially to a plateau at 50--70\% progress, matching real RLHF convergence.
Ground-truth hacking labels and peak progress are attached to each job but hidden from the scheduler via a whitelist-based proxy (\texttt{SchedulerJobView}), so schedulers observe only the current loss, current eval score, and historical eval checkpoints.

\begin{table}[h]
\caption{Simulator hyperparameters and job-type profiles.}
\label{tab:sim-params}
\centering
\footnotesize
\setlength{\tabcolsep}{4pt}
\begin{tabular}{@{}lc@{}}
\toprule
\textbf{Parameter} & \textbf{Value} \\
\midrule
\multicolumn{2}{@{}l}{\emph{Cluster / Workload}} \\
Total GPUs & 32 (E1) / 64 (E2--E6) \\
Preemption overhead & 2.0 min \\
Jobs / Tenants / Seeds & 200 / 5 / 5 \\
Arrival process & Poisson, $\lambda{=}1.0$ jobs/min \\
\midrule
\multicolumn{2}{@{}l}{\emph{LoRA jobs}} \\
Duration / GPUs / Eval freq & 10--60\,min / 1--2 / 10\% prog. \\
Eval curve & Monotonic (exponential) \\
\midrule
\multicolumn{2}{@{}l}{\emph{DPO jobs}} \\
Duration / GPUs / Eval freq & 30--120\,min / 2--4 / 20\% prog. \\
Eval curve & Diminishing returns \\
\midrule
\multicolumn{2}{@{}l}{\emph{RLHF jobs}} \\
Duration / GPUs / Eval freq & 60--360\,min / 4--8 / 15\% prog. \\
Reward-hack fraction & 60\% hacking, 40\% healthy \\
Loss plateau & $\mathcal{U}(0.50, 0.70)$ progress \\
Hacking peak / drop & $\mathcal{U}(0.55,0.75)$ / $\mathcal{U}(0.10,0.30)$ \\
Healthy eval curve & Monotone concave + noise \\
\midrule
\multicolumn{2}{@{}l}{\emph{Evaluation}} \\
Quality $\tau$ / min improvement & 0.3 / 1\% relative \\
Eval duration & 1 / 3 / 5 min (small / med / large) \\
\bottomrule
\end{tabular}
\end{table}

\section{E1: Mixed Workload Full Results}
\label{app:e1}

Table~\ref{tab:e1-full} presents the complete E1 results (50\% LoRA, 30\% DPO, 20\% RLHF, 32 GPUs) omitted from the main text for space.
EvalStop+SRTF achieves the best JCT and competitive TTFUC with no fairness penalty.

\begin{table}[h]
\caption{E1: Mixed workload (50/30/20 LoRA/DPO/RLHF, 32 GPUs, 200 jobs, 5 seeds). Precision / FPR columns only defined for stop-enabled schedulers; ``n/a'' = no stops issued.}
\label{tab:e1-full}
\centering
\resizebox{\columnwidth}{!}{%
\begin{tabular}{@{}lcccccc@{}}
\toprule
Scheduler & JCT & TTFUC & Wasted & Saved & Prec. & FPR \\
\midrule
FIFO          & 450{\scriptsize$\pm$57} & 381{\scriptsize$\pm$54} & 34.2\% & 0.0\%  & n/a    & 0.0\% \\
SJF-Est       & 406{\scriptsize$\pm$57} & 336{\scriptsize$\pm$54} & 34.7\% & 0.0\%  & n/a    & 0.0\% \\
SRTF-Est      & 361{\scriptsize$\pm$39} & 290{\scriptsize$\pm$36} & 33.9\% & 0.0\%  & n/a    & 0.0\% \\
LossAware     & 722{\scriptsize$\pm$51} & 273{\scriptsize$\pm$39} & 34.5\% & 0.0\%  & n/a    & 0.0\% \\
EvalSched     & 1061{\scriptsize$\pm$107} & 548{\scriptsize$\pm$34} & 33.4\% & 0.0\% & n/a    & 0.0\% \\
StopAt0.5+SRTF  & \textbf{290}{\scriptsize$\pm$26} & 240{\scriptsize$\pm$25} & 21.0\% & 33.3\% & 58.9\% & 9.7\% \\
StopAt0.65+SRTF & 312{\scriptsize$\pm$29} & 256{\scriptsize$\pm$28} & 24.3\% & 22.5\% & 58.9\% & 9.7\% \\
LossPlateau+SRTF & 354{\scriptsize$\pm$48} & 287{\scriptsize$\pm$45} & 32.0\% & 3.6\%  & 30.8\% & 13.1\% \\
\midrule
EvalStop+SRTF & 344{\scriptsize$\pm$35} & 279{\scriptsize$\pm$32} & \textbf{28.5\%} & 8.5\% & \textbf{100\%} & \textbf{0.0\%} \\
\bottomrule
\end{tabular}%
}
\end{table}

\section{E5--E6: Numerical Details}
\label{app:e5e6}

Tables~\ref{tab:e5-full} and~\ref{tab:e6-full} provide numerical values for the threshold and eval-frequency sensitivity experiments plotted in Figures~\ref{fig:threshold} and~\ref{fig:evalfreq}.

\begin{table}[h]
\caption{E5: Threshold sensitivity (EvalStop+SRTF, 80\% RLHF, 64 GPUs, 5 seeds). Row with $k{=}2$ (our default) is highlighted.}
\label{tab:e5-full}
\centering
\small
\begin{tabular}{@{}ccccc@{}}
\toprule
$k$ & JCT & TTFUC & Wasted & Stopped \\
\midrule
1 & 874{\scriptsize$\pm$50} & 761{\scriptsize$\pm$47} & 19.4\% & 127 \\
\textbf{2} & \textbf{1018}{\scriptsize$\pm$53} & \textbf{883}{\scriptsize$\pm$50} & \textbf{26.9\%} & \textbf{93} \\
3 & 1093{\scriptsize$\pm$56} & 946{\scriptsize$\pm$54} & 32.8\% & 57 \\
4 & 1123{\scriptsize$\pm$61} & 971{\scriptsize$\pm$58} & 34.6\% & 2 \\
5 & 1123{\scriptsize$\pm$59} & 971{\scriptsize$\pm$56} & 34.5\% & 0 \\
\bottomrule
\end{tabular}
\end{table}

\begin{table}[h]
\caption{E6: Eval frequency sensitivity (EvalStop+SRTF, 80\% RLHF, 64 GPUs, 5 seeds). $\Delta$JCT and $\Delta$Wasted are reported vs.\ SRTF-Est alone (JCT${=}$1124, Wasted${=}$34.5\%).}
\label{tab:e6-full}
\centering
\small
\begin{tabular}{@{}ccccc@{}}
\toprule
Eval freq & JCT & Wasted & Saved & $\Delta$JCT vs SRTF \\
\midrule
5\%  & 867{\scriptsize$\pm$75}  & 17.6\% & 24.1\% & $+$22.8\% \\
10\% & 993{\scriptsize$\pm$34}  & 23.2\% & 14.8\% & $+$11.6\% \\
\textbf{15\%} & \textbf{1018}{\scriptsize$\pm$53} & \textbf{26.9\%} & \textbf{9.9\%} & \textbf{$+$9.4\%} \\
20\% & 1111{\scriptsize$\pm$96} & 34.2\% & 4.8\% & $+$1.1\% \\
30\% & 1123{\scriptsize$\pm$63} & 29.8\% & 0.7\% & $+$0.1\% \\
\bottomrule
\end{tabular}
\end{table}

\end{document}